\documentclass[preprint,12pt,authoryear]{elsarticle}

\usepackage{amssymb}

\usepackage[utf8]{inputenc}
\usepackage[margin=1in]{geometry}
\usepackage{natbib}
\usepackage{enumitem}
\usepackage{amsmath,amssymb}
\usepackage{makecell}
\usepackage{adjustbox}
\usepackage{graphicx}
\usepackage{float}
\usepackage{xcolor}
\usepackage{booktabs}
\usepackage{setspace}
\usepackage{comment}
\usepackage{amsfonts}
\usepackage{units}
\usepackage[ruled,vlined]{algorithm2e}
\usepackage[citecolor=blue,colorlinks=true,linkcolor=blue]{hyperref}
\usepackage[english]{babel}
\usepackage{amsthm}
\usepackage{dsfont}

\theoremstyle{definition}
\newtheorem{definition}{Definition}[section]
\newtheorem{assumption}{Assumption}[section]

\DeclareMathOperator*{\argmax}{arg\,max}
\DeclareMathOperator*{\argmin}{arg\,min}

\journal{Neurocomputing}

\begin{document}

\begin{frontmatter}



\title{Enhanced method for reinforcement learning based dynamic obstacle avoidance by assessment of collision risk}


\author[label1]{Fabian Hart}
\author[label1]{Ostap Okhrin}
\address[label1]{Chair of Econometrics and Statistics, esp. in the Transport Sector, Technische Universität Dresden, Dresden,
            01062, Germany}

\begin{abstract}
In the field of autonomous robots, reinforcement learning (RL) is an increasingly used method to solve the task of dynamic obstacle avoidance for mobile robots, autonomous ships, and drones. 
A common practice to train those agents is to use a training environment with random initialization of agent and obstacles. 
Such approaches might suffer from a
low coverage of high-risk scenarios in training, leading to impaired final performance of obstacle avoidance.
This paper proposes a general training environment where we gain control over the difficulty of the obstacle avoidance task by using short training episodes and assessing the difficulty by two metrics: The number of obstacles and a collision risk metric. 
We found that shifting the training towards a greater task difficulty can massively increase the final performance.
A baseline agent, using a traditional training environment based on random initialization of agent and obstacles and longer training episodes, leads to a significantly weaker performance. 
To prove the generalizability of the proposed approach, we designed two realistic use cases: A mobile robot and a maritime ship under the threat of approaching obstacles. 
In both applications, the previous results can be confirmed, which emphasizes the general usability of the proposed approach, detached from a specific application context and independent of the agent's dynamics. 
We further added Gaussian noise to the sensor signals, resulting in only a marginal degradation of performance and thus indicating solid robustness of the trained agent.
\end{abstract}


\end{frontmatter}


\section{Introduction}
Nowadays, autonomous robots have a wide range of applications, such as mobile robots in industry or hospitals, drones, autonomous underwater vehicles, etc. The fundamental task of autonomous real-world navigation is to reach a global goal (global navigation) while maneuvering in a dynamic environment where it is necessary to react to static and dynamic obstacles (local obstacle avoidance) \citep{patle2019review}. With the rise of machine learning techniques, reinforcement learning (RL) has widely been studied in the field of autonomous robotics \citep{singh2021reinforcement}. Instead of using supervised learning techniques to imitate some prerecorded behavior, which can still be sub-optimal, RL trains agents according to a predefined reward function. An often applied strategy in RL-based autonomous navigation is to separate the overall goal of global navigation and local obstacle avoidance since efficiently reaching a global goal usually stands in contrast to ensuring safety under the threat of obstacles \citep{wang2018learning, ding2018hierarchical}. This work focuses solely on local obstacle avoidance and particularly on dynamic obstacles due to its complexity for safe navigation. 

In RL, an agent learns by interacting with a training environment, and therefore the definition of this environment strongly impacts the final policy. A common approach for RL-based dynamic obstacle avoidance is to define a training environment where dynamic obstacles, particularly their position, speed, and direction, are randomly initialized all over the environment \citep{arvind2019autonomous, wang2018learning, xu2022path, yuan2021auv, fan2020distributed, ding2018hierarchical, kastner2021arena}. Unfortunately, in some studies, we fail to find information about how a training episode is defined or how exactly the dynamic obstacles are initialized \citep{qiao2008application, tong2021uav}. We claim that by using traditional approaches of randomly moving obstacles, the coverage of situations with high collision risk in training is relatively low, leading to a final policy that is just suitable for simple collision avoidance tasks. Such high-risk situations can be, for example, characterized as situations where simultaneously multiple obstacles pose a threat to the agent or where just a small portion of steering maneuvers are possible to avoid collisions.

Some approaches use at least simple heuristics to increase the collision risk in training: While initializing a training episode, one obstacle (among multiple) is being created "to pose a threat" to the safety of the agent \citep{xu2022path}, however, we miss a clear definition of how this is done. \cite{chun2021deep} use obstacles that are initialized randomly but in a particular range and with a particular moving direction to increase the threat for the agent. \cite{lan2020cooperative, xie2020composite} define movement trends of dynamic obstacles that intersect with that of the agent in a way that collisions are inevitable without avoidance strategy. \cite{li2021path} define one dynamic obstacle that is forming a threat to the planned path of the agent. 
From our point of view, the mentioned approaches are mainly based on the designer's intuition to increase the collision risk in training, missing a formal definition of collision risk.

To summarize, most of the current studies on RL-based dynamic obstacle avoidance use a completely random initialization of their training environment, and there are just a few who use some heuristics to increase the collision risk in training. 
To fill this gap, our contribution is as follows:
\begin{itemize}
    \item We propose a training environment where the difficulty of the obstacle avoidance task can be adjusted via two metrics: (i) a collision risk metric that aims to quantify the risk of a collision considering the agent's dynamics and (ii) the number of simultaneous obstacles that pose a threat to the agent. 

    \item We perform an analysis that investigates how the difficulty in training affects the final policy, and we demonstrate that shifting training towards greater difficulty can massively increase the final performance of the obstacle avoidance task. 
    
    \item For comparison, we implement a traditional training environment for dynamic obstacle avoidance based on random initialization of obstacles. The results indicate that our training method significantly outperforms traditional approaches to designing a training environment.

   \item To demonstrate the generalizability, we transfer our proposed approach to two further application domains: Dynamic obstacle avoidance for a mobile robot and an autonomous ship. The results indicate that our approach is suitable for training autonomous agents from different obstacle avoidance domains regardless of their individual maneuverability. Furthermore, we prove the trained agent's robustness by adding Gaussian noise to the sensor signals. 
\end{itemize}
Bringing together, this study gives general recommendations for how to define a training environment for dynamic obstacle avoidance by increasing the difficulty of obstacle avoidance tasks in training. Instead of using rule-of-thumb heuristics to increase the difficulty, we provide a standardized approach that uses two metrics to quantify the difficulty of the obstacle avoidance task. We emphasize that, due to this standardization, our approach is not restricted to a specific application but can rather be applied in all domains of obstacle avoidance. 

This work is structured as follows: In Section \ref{sec:problem}, we give a formal definition of the problem, followed by a definition of a collision risk metric in Section \ref{sec:CRmetric}. We further describe the RL methodology in Section \ref{sec:RLmethodology} and define a training environment for dynamic obstacle avoidance in Section \ref{sec:training} with its training results detailed in Section \ref{sec:results}. In Section \ref{sec:generelizability}, we investigate the generalizability of the proposed training approach, and we conclude in Section~\ref{sec:conclusion}.

\section{Problem statement}\label{sec:problem}
This study investigates the capability of successful obstacle avoidance for an agent trained with reinforcement learning (RL). For reasons of simplification, we construct a general, two-dimensional use case detached from a specific transportation or robot context, where obstacles follow a linear trajectory. As motivated in the introduction, we are solely focusing on obstacle avoidance without providing a global path to follow, therefore the task of the agent is to just navigate between obstacles without collision. There are two different control inputs: A longitudinal force and a torque. This configuration is similar to many real applications, such as (i) autonomous mobile robots with a longitudinal acceleration and a lateral steering input \citep{alatise2020review}, (ii) autonomous ships with a longitudinal thrust and a rudder angle input which can be translated into a torque \citep{munim2019autonomous}, or (iii) plane-like drones with a longitudinal force via engines or propellers and a rudder input \citep{floreano2015science}. We assume a two-dimensional space, resulting in a three-degree-of-freedom (3-DoF) model, however, the proposed space can be straightforwardly generalized to higher dimensions. The state of such a model is defined via the position vector with heading angle $\eta = (x, y, \psi)^\top \in	\mathbb{R}^3 $ and the velocity vector $\nu = (u, v, r)^\top  \in	\mathbb{R}^3$ containing velocities in $x$- and $y$-direction and the yaw rate $r$. Based on these state vectors, the dynamics of a 3-DoF model can generally be expressed via two differential equations: 
\begin{align}
    \Dot{\eta} &= g(\nu),  \label{eq:eta_dot}\\
    \Dot{\nu} &= f(\nu, \tau), \label{eq:nu_dot}
\end{align}
with the control vector  $\tau = (\tau_u, \tau_v, \tau_r)^\top  \in	\mathbb{R}^3 $ and where $f(\cdot,\cdot)$ and $g(\cdot)$ are deterministic functions specifying dynamics. Through this work, we define three different agent dynamics: At first, we investigate the problem of dynamic obstacle avoidance with a general agent, and, in Section \ref{sec:generelizability}, we will define two more realistic agent dynamics. 

The initial agent is a simple, point-mass agent where the longitudinal force ($\tau_u$) and the torque ($\tau_r$) are controlled by the RL agent. This translates equation (\ref{eq:nu_dot}) and (\ref{eq:eta_dot}) to:

\begin{equation}\label{eq:agent_dyn}
\begin{aligned}[c]
    \dot{x} &= u\cos(\psi),\\
    \dot{y} &= u\sin(\psi),\\
    \dot{\psi} &= r,
\end{aligned}
\quad\quad\quad
\begin{aligned}[c]
    \dot{u} &= \tau_u/m,\\
    \dot{v} &= 0,\\
    \dot{r} &= \tau_r/I,
\end{aligned}
\end{equation}
with agent mass $m = \unit[15]{kg}$ and moment of inertia $I = \unit[30]{kg m^2}$.

In accordance with the definition of the agent, an obstacle $i$ for $i=1, \ldots ,N_{\rm obst}$ is defined via the position vector $\eta_{\rm obst}^i$ and the velocity vector $\nu_{\rm obst}^i$. For the sake of simplicity, we just consider obstacles that follow a linear trajectory which leads to $\nu_{\rm obst}^i = (u_{\rm obst}^i, 0, 0)^\top $ with $u_{\rm obst}^i$ being constant over time.
Furthermore, we define a collision area with radius $R_{\rm coll} = \unit[3]{m}$ around agent and obstacles, and an overlap of both areas is defined as a collision. Some of the variables are depicted in Figure \ref{fig:technical} later on.

\section{Collision risk metric}\label{sec:CRmetric}
As motivated in the introduction, we aim for a dynamic obstacle training environment that increases the difficulty of the obstacle avoidance task by using a collision risk metric. For this study, we adopt the general definition of risk based on probability concepts from \cite{andretta2014some}: 
\begin{definition}
Risk is the probability of an unwanted event.
\end{definition}
This definition is pretty natural and lies in $[0;1]$ because of the properties of probabilities.
In this study, an unwanted event refers to the collision of an agent with an obstacle. This leads to the definition:
\begin{definition}\label{def:CR}
Collision risk is defined by the probability of a future collision with an obstacle.
\end{definition}
It is worth mentioning that it is not in the scope of this study to propose the best possible collision risk metric but more how the performance of obstacle avoidance can be increased by using a common and known collision risk metric in training.
A list of different metrics to quantify the risk of a collision has already been proposed in the literature. Most related studies focus on maritime traffic, where risk analysis is an essential step in conducting ship avoidance maneuvers. Various studies use the closest point of approach (CPA) as a basis. The CPA defines the point where two ships come closest under the assumption of constant velocities. Time to the closest point of approach (TCPA) and distance to the closest point of approach (DCPA) are common indicators for collision risk in various studies, such as \cite{zhen2017novel} or \cite{liu2019novel}. Another approach for collision risk analysis is to formulate an area around a ship that should be kept clear to avoid a collision \citep{szlapczynski2018ship}.  

 In this work, we adopt the concept of collision risk from \cite{huang2020time} who defined a collision risk metric as \emph{the fraction of all possible maneuvers that lead to a collision}, following Definition \ref{def:CR}.
 An advantage of their approach over others is the consideration of the agent's maneuverability to compute the collision risk. This is important when, for example, an agent with good maneuverability might be able to avoid a collision with a close by obstacle while an agent with poor maneuverability might not, assuming the same scenario. Formally, the collision risk according to \cite{huang2020time} from the law of total probabilities is:
\begin{equation*}
    CR = P(collision) = \sum_{i=1}^n P(collision|\tau_i)P(\tau_i),
\end{equation*}
where $\tau_i$ is a control command, $P(\tau_i)$ is the probability to choose $\tau_i$, $P(collision|\tau_i)$ is the probability to collide when choosing $\tau_i$, and $n$ is the number of all possible control commands. Following \cite{huang2020time}, the following assumption is made:
\begin{assumption}
The probability of a maneuver $P(\tau_i)$ yields a uniform distribution.
\end{assumption}
This means that all possible maneuvers are equally likely to be chosen without any prior knowledge.
For the computation of the collision probability $P(collision|\tau_i)$, the \emph{Generalized Velocity Obstacles} (GVO) approach is used, introduced by \cite{wilkie2009generalized}. GVO is a common method for obstacle avoidance and has widely been applied, such as in \cite{mahmoodi2014real}, \cite{peterson2015virtual}, \cite{huang2019generalized}. Following this approach, the occurrence of future collision can be analytically calculated under the following simplifying assumptions \citep{huang2020time}:
\begin{assumption}
Agent and obstacles are disk-shaped.
\end{assumption}
\begin{assumption}
The obstacles' trajectories are known and linear (the obstacles follow their initial velocity vector).
\end{assumption}
\begin{assumption}
Only those collisions are addressed that will happen before a finite time horizon $t_{\rm lim}$.
\end{assumption}
\begin{assumption}
The control command $\tau_i$ remains constant within $t_{\rm lim}$.
\end{assumption}
From our perspective, especially the last assumption impairs the suitability of this approach to give a good estimate of the current collision risk: It would be more realistic for the agent to be able to react to a threat by choosing another control input within the time horizon $t_{\rm lim}$. To overcome this issue, we plan to develop our own collision risk metric in the future.

\begin{figure}[!ht]
    \centering
    \includegraphics[width=1\linewidth]{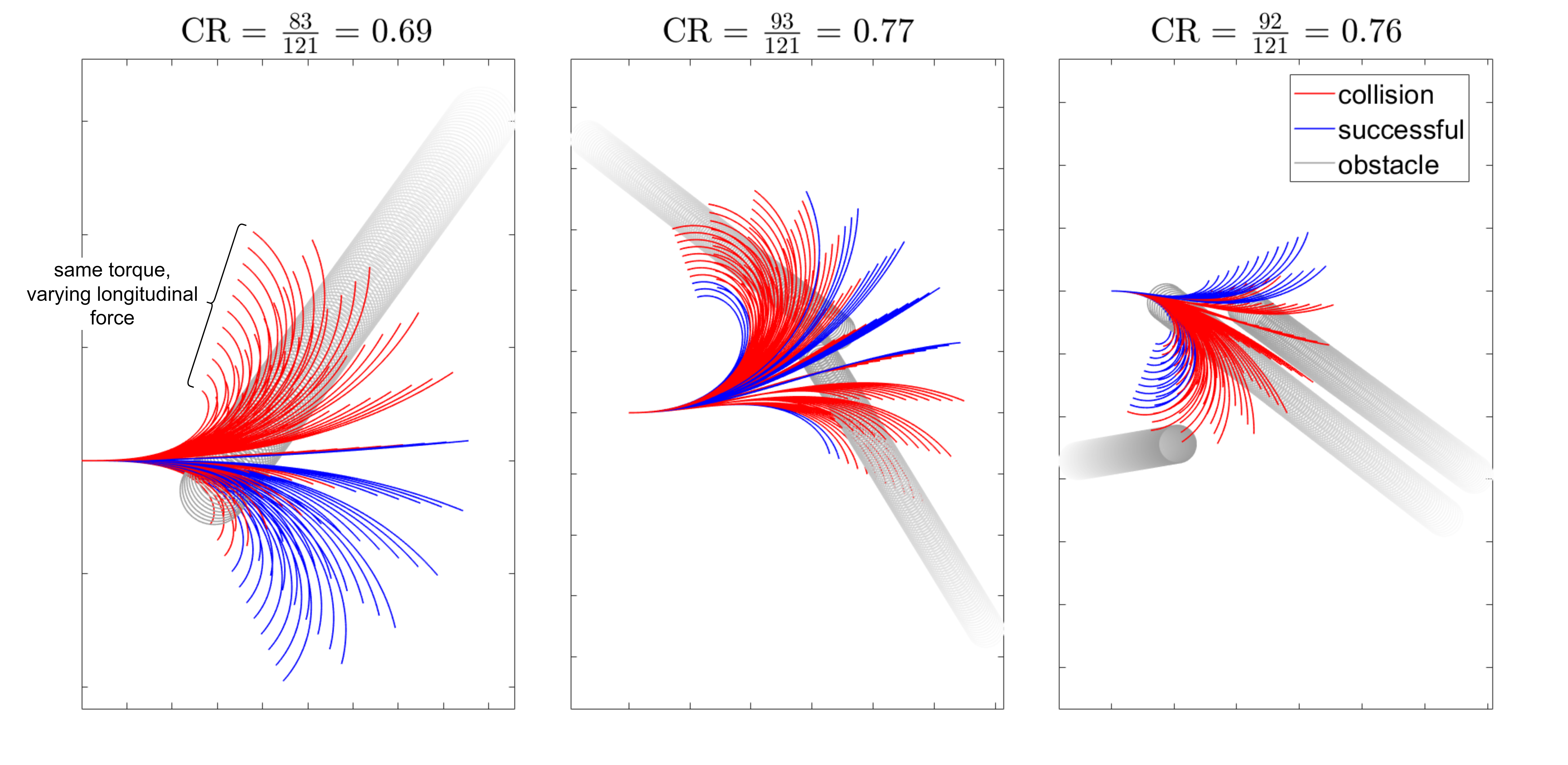}
    \caption{Agent trajectories for different, constant control inputs ($\tau)$. Trajectories leading to a collision with an obstacle trajectory (grey) are marked with red, successful ones with blue color. The resulting initial collision risk $CR$ is displayed in the caption of each scenario. }
    \label{fig:CR}
\end{figure}

Figure \ref{fig:CR} illustrates the concept of collision risk with the help of three exemplary scenarios. The obstacle with its collision area for different time points $R_{\rm coll}$ is represented in grey. The agent's possible trajectories under constant control input $\tau$ and within the time horizon $t_{\rm lim}$ are depicted in blue and red. For reasons of readability, the collision area of the agent is not drawn in. For this illustration, we considered $11$ equally spaced values for each control command, longitudinal force $\tau_u$, and torque $\tau_r$, resulting in 121 different trajectories. This configuration is kept unchanged for the entire study. Blue trajectories represent collision-free trajectories, while red ones represent trajectories that lead to a collision with one of the obstacles. The collision risk can now be simply computed by dividing the number of collision trajectories by the number of all possible trajectories. The final CR value is displayed above each scenario. 

\section{Reinforcement learning methodology}
\label{sec:RLmethodology}
RL is a sub-field of machine learning and has widely been used to develop self-learning agents. In the RL framework, an RL agent learns by interacting with the environment and thus is able to optimize sequential decision-making problems. In this study, the RL agent represents an autonomous robot that has to make decisions about what maneuver to execute in order to avoid dynamic obstacles. 
\subsection{Basics}
 RL is grounded on Markov decision processes that are defined by the following components \citep{sutton2018reinforcement}: A state space $\mathcal{S}$ that, in our case, includes state variables of agent and obstacles; an action space $\mathcal{A}$ that includes longitudinal force and torque; an initial state distribution $T_0: \mathcal{S} \rightarrow [0,1]$ that defines the initial position and velocities of agent and obstacles; a state transition probability distribution $\mathcal{P}: \mathcal{S} \times \mathcal{A} \times \mathcal{S} \rightarrow [0,1]$ that defines the movement of agent and obstacles; a reward function $\mathcal{R}: \mathcal{S} \times \mathcal{A} \rightarrow \mathbb{R}$ that should incentives the agent to avoid collisions; and a discount factor $\gamma \in [0,1]$. In an iterative fashion, the agent receives a state information $s_t \in \mathcal{S}$ at each time step $t$, then it selects a two-dimensional action $a_t \in \mathcal{A}$, gets an instantaneous reward $r_{t+1}$, and transits to the next state $s_{t+1} \in \mathcal{S}$. Formally, a capital notation is used to describe random variables, e.g., $S_t$, and a small notation, e.g., $s_t$ or $s$, to describe realizations of random variables.

The objective of RL is to maximize the expected discounted cumulative reward when starting from state  $S_0$: $J(\pi) =  E_{\pi}\left[ \sum_{k=0}^{\infty} \gamma^k R_{k+1}\right | S_0 = s]$. Therefore, the RL agent has to learn a policy $\pi$ that maps from states $s \in \mathcal{S}$ to actions $ a \in  \mathcal{A}$.
 In value-based RL, the action-value function $Q^{\pi}(s,a)$ is used to assess the quality of a state-action pair. Formally, the action-value is defined as the expected return when starting in state $s$, taking action $a$, and following policy $\pi$ afterward: $Q^{\pi}(s,a) = E_{\pi}\left[ \sum_{k=0}^{\infty} \gamma^k R_{t+k+1} | S_t = s, A_t = a \right]$.  Furthermore, we assume the existence of an optimal action-value function $Q^*(s,a) = \max_{\pi} Q^{\pi}(s,a)$. Accordingly, there is a deterministic optimal policy $\pi^*(s) = \argmax_{a \in \mathcal{A}} Q^*(s,a)$ that we try to find. To approximate $Q^*(s,a)$, the  \cite{bellman1954theory} optimality equation is a fundamental practice: 
\begin{equation*}
    Q^*(s,a) = \mathcal{R}(s,a) + \gamma \sum_{s' \in \mathcal{S}} \mathcal{P}_{sa}^{s'} \max_{a' \in \mathcal{A}} Q^*(s',a').
\end{equation*}

An important algorithm that uses the Bellman Equation is the $Q$-learning algorithm \citep{watkins1992q}, where a finite number of state-action pairs are stored as a tabular representation. However, this allows just for discrete state and action spaces. An extension of $Q$-learning for continuous state spaces is the deep $Q$-network (DQN) algorithm \citep{mnih2015human} where $Q$-values are approximated by more complex functions instead of finite tables. Commonly used function approximators are neural networks. To train the function $Q^{\omega}(s,a)$ with parameter vector $\omega$, the gradient descend algorithm is applied:
\begin{align*}
    \omega & \leftarrow \omega + \alpha \left\{y-Q^{\omega}(s,a) \right\} \nabla_{\omega} Q^{\omega}(s,a), \\
    y &= r + \gamma \max_{a' \in \mathcal{A}} Q^{\omega'}(s',a'),
\end{align*}
with learning rate $\alpha$ and $Q^{\omega'}(s,a)$ as target network. The target network represents a copy of the original network, but it is held constant for some fixed number of time steps in order to stabilize the training process. Another extension of DQN is experience replay, which enables the agent to memorize past experiences that are used for learning. These experiences are sampled randomly from the replay buffer for training in a mini-batch scheme. One drawback coming along with the DQN algorithm is the restriction to discrete action spaces. However, our use case requires continuous control commands. We, therefore, use the twin delayed deep deterministic policy gradient (TD3) algorithm, introduced by \cite{fujimoto2018addressing}.

\subsection{Twin delayed deep deterministic policy gradient (TD3)}

TD3 uses several features of the deep deterministic policy gradient (DDPG) algorithm \citep{lillicrap2015continuous}. DDPG is a dual-network algorithm that incorporates a $Q$-function (Critic) $Q^{\omega}(s,a)$, just as the DQN algorithm, and a deterministic policy-function (Actor) $\mu^{\theta}: \mathcal{S} \rightarrow \mathcal{A}$ with parameter vector $\theta$. Both functions are usually represented as neural networks. \cite{lillicrap2015continuous} adopted experience replay and target networks from DQN but introduced a soft update of the target networks:
\begin{align}\label{eq:DDPG_soft_tgt_up}
    \omega' &= \tau \omega + (1-\tau) \omega', \nonumber \\ 
    \theta' &= \tau \theta + (1-\tau) \theta',
\end{align}
with hyperparameter $\tau$ as the soft target update rate and $ \omega'$ and $\theta'$ as parameters of the target networks. For small values of $\tau$, the target networks update smoothly, which has been shown to improve the training stability. Furthermore, exploration is enhanced by adding random noise to the actions. However, there are still some limitations of the DDPG algorithm. A main issue is the overestimation in $Q$-values which can lead to a sub-optimal policy \citep{dong2020deep,waltz2022two}. Another problem is the sensitivity to the choice of hyperparameters which results in impaired training stability. To overcome these limitations, \cite{fujimoto2018addressing} introduced the TD3 algorithm. The first modification over DDPG is the incorporation of double $Q$-learning where the minimum of two $Q$-functions is selected as the target $Q$-value: 
\begin{equation*}
     y = r + \min_{j=1,2} Q^{\omega'_j}(s,a).
\end{equation*}
Using the minimum of both $Q$-functions has been found to combat the overestimation of $Q$-values. A further modification in TD3 is the addition of a truncated normal distribution noise to each action: 
\begin{equation}
    \Tilde{a}_{i+1} = \mu^{\theta'}(s_{i+1}) + \Tilde{\epsilon}, \quad \text{with } \Tilde{\epsilon} \sim \text{clip}\{\mathcal{N}(0, \Tilde{\sigma}),-c,c\},
\end{equation}
for some $c > 0$. This regularization method is applied to reduce the variance of the critic update and to smooth the computation of $Q$-values. To further improve the convergence performance, \cite{fujimoto2018addressing} proposed updating policy and target networks less frequently. A typical strategy is to perform the update every $d = 2$ steps. The complete algorithm is detailed in Algorithm \ref{algo:TD3}.

\subsection{Architecture and hyperparameters}
Actor $\mu^{\theta}$ and critic function $Q^{\omega}(s,a)$ are feed-forward neural networks with two layers of hidden neurons and 128 neurons in each layer. All layers use ReLU activation functions (\cite{relu}), except the output layer of the actor network where a $\rm tanh(\cdot)$ activation function is applied. The learning rates $\alpha_{\rm actor}$ and $\alpha_{\rm critic}$ that are used to update actor and critic are set to 0.001. Optimization is performed with Adam \citep{kingma2014adam}. Table \ref{tab:hyperparams} displays the complete list of hyperparameters.
\begin{table}[H]
    \centering
    \begin{tabular}{l|l}
    Hyperparameter & Value\\
    \toprule
        Discount factor $\gamma$ &  0.99 \\
        Batch size $N$ & 32\\
        Replay buffer size $|\mathcal{D}|$ & $10^5$ \\
        Learning rate actor $\alpha_{\rm actor}$ & $0.0001$ \\ 
        Learning rate critic $\alpha_{\rm critic}$ & $0.0001$ \\ 
        Soft target update rate $\tau$ & 0.001 \\
        Optimizer & Adam \\
        Target noise $\sigma$ & 0.2 \\
        Target noise clip $c$ & 0.5 \\
       	Number of hidden layers & 2 \\
		Neurons per hidden layer & 128 \\
		policy update frequency $d$ & 2\\ 
    \end{tabular}
    \caption{List of TD3 hyperparameters. }
    \label{tab:hyperparams}
\end{table}

\section{Training} \label{sec:training}
\subsection{State, action and reward} \label{sec:state_action_reward}
\begin{figure}[ht]
    \centering
    \includegraphics[width=0.7\linewidth]{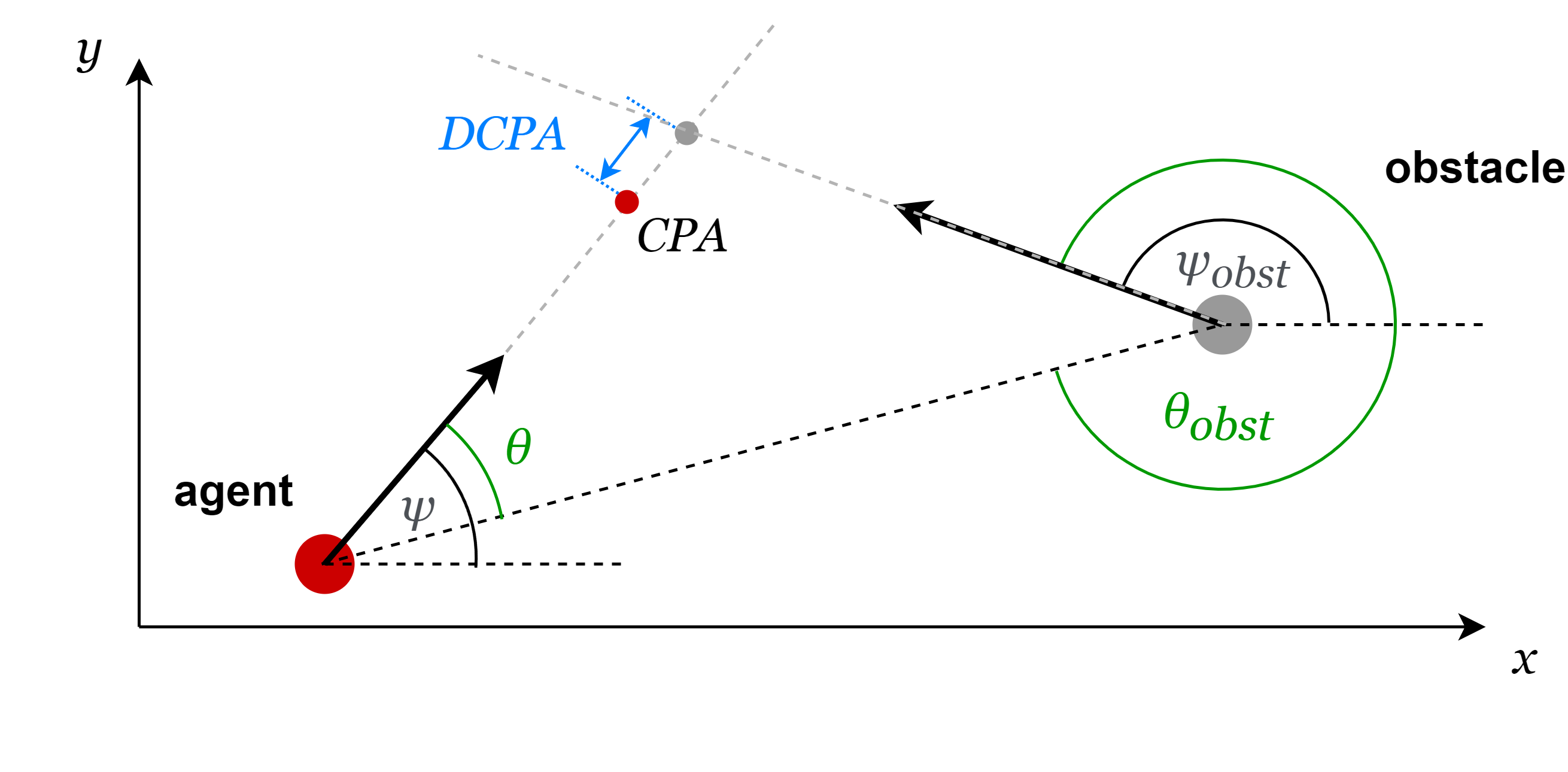}
    \caption{Technical illustration of the agent and an exemplary obstacle with the closest point of approach (CPA) and the distance to the closest point of approach (DCPA).}
    \label{fig:technical}
\end{figure}

Although we used different agent dynamics throughout this work, we aim to define a generic action space, state space, and reward that is kept constant for all agent dynamics. Since we consider an agent to be controlled via a longitudinal force and a torque,  we configured a two-dimensional action $a_t = (a_{u,t}, a_{r,t})^\top \in [-1, 1]^2$ that is mapped to the control vector $\tau_t$ at each time step $t$:
\begin{align*}
    \tau_t &= \tau_{\rm max} a_t, \\
    \tau_{\rm max} &= \begin{pmatrix}
            \tau_{u,\rm max} & 0 \\
            0 & 0\\
            0 & \tau_{r,\rm max} \\
	\end{pmatrix} ,
\end{align*}
with the maximum longitudinal force $\tau_{u,\rm max}$ and the maximum torque $\tau_{r,max}$. To make adequate decisions, the agent receives information about itself and surrounding obstacles at each time step $t$, resulting in the state $s_t$, where scaling parameters are used to bring all state variables in approximately the same range:
\begin{align}
    s_t &= \begin{pmatrix}
            \frac{u_t}{u_{\rm scale}}, & \frac{r_t}{r_{\rm scale}}, & \frac{d_t^i}{d_{\rm scale}}, & \frac{u_{\rm obst,t}^i}{u_{\rm scale}}, & \frac{\theta_t^i}{\pi}, & \frac{\theta_{\rm obst,t}^i}{\pi}, & \frac{DCPA_t^i}{d_{\rm scale}}, & \frac{TCPA_t^i}{t_{\rm scale}}\label{eq:state_space}\\
	\end{pmatrix}^\top  ,
\end{align}
with an Euclidean distance to the obstacle $i$ being:
\begin{align*}
    d_t^i &= \sqrt{(x_t - x_{\rm obst, t}^i)^2 + (y_t - y_{\rm obst, t}^i)^2 }.
\end{align*}
The distance to obstacle $i$ at the closest point of approach $DCPA_t^i$ under control command $\tau = 0$ is defined as:
\begin{align*}
    DCPA_t^i &= \min_{t'} (d_{t'}^i|\tau = 0, t' > t),
\end{align*}
and $TCPA_t^i$ is the time until the closest point of approach for obstacle $i$ is reached:
\begin{align*}
    TCPA_t^i &= \argmin_{t'} (d_{t'}^i|\tau = 0, t' > t).
\end{align*}
The angles $\theta_{t}^i$ and $\theta_{\rm obst,t}^i$, as well as the closest point of approach $CPA$ and the distance to the closest point of approach $DCPA$, are illustrated in Figure \ref{fig:technical}. The state variables $DCPA^i_t$ and $TCPA^i_t$ are used to provide the agent with more information about the criticality of an obstacle $i$ at time point $t$. These additional state variables have been shown to enhance the final performance of obstacle avoidance. The first two variables of the state vector are agent related, while the other six contain information about surroundings. Consequently, the size of the state vector $s_t$ is $N = 2 + 6 N_{\rm obst}$. The obstacles are sorted into the state vector $s_t$ with ascending $TCPA$. All environment parameters with corresponding values can be found in Table \ref{tab:env_parameter}.

\begin{table}[H]
    \centering
    \begin{tabular}{l|l}
   Parameter & Value \\
    \toprule
    $\tau_{u, \max}$  & $\unit[1]{N}$\\
    $\tau_{r, \max}$  & $\unit[1]{Nm}$\\
    $u_{\rm scale}$ & $\unit[1.5]{m/s}$\\
    $r_{\rm scale}$ & $\unit[0.5]{s^{-1}}$\\
    $d_{\rm scale}$ & $\unit[15]{m}$\\
    $t_{\rm scale}$ & $\unit[20]{s}$\\
    $\Delta t$ & $\unit[0.1]{s}$\\
    \end{tabular}
    \caption{Parameters of the environment.}
    \label{tab:env_parameter}
\end{table}

Following RL literature, we aim to keep the reward function as simple as possible \citep{silver2018general}. Therefore, we impose a non-zero reward just at the end of an episode $t_{\rm end}$:
\begin{equation} 
\label{eq:reward}
	r_t = 
	\begin{cases}
	     0, & \text{for } t<t_{\rm end},\\
		+1, & \text{for } t=t_{\rm end} \text{ and episode is successful},\\
		-1, & \text{for collision},\\
	\end{cases}
\end{equation}
where a collision is defined as an overlap between the collision area $R_{\rm coll}$ of an obstacle and the agent. The episode ends in case of a collision. 

\subsection{CR-based training environment}\label{sec:training_env}
As motivated in the introduction, we aim to enhance the final performance of obstacle avoidance by increasing task difficulty in training. We quantify the difficulty via the combination of two metrics: First, we use the collision risk metric by \cite{huang2020time}, described in Section \ref{sec:CRmetric}. However, as stated above, our method is not limited to a specific collision risk metric. And as a second metric for the difficulty in training, we use the number of obstacles that pose a threat to the agent. 

In many studies on RL-based obstacle avoidance, one training episode is quite long and contains encounters with multiple obstacles in a sequential fashion \citep{wang2018learning, xu2022path, arvind2019autonomous, yuan2021auv, chun2021deep, fan2020distributed, ding2018hierarchical, huang2018reinforcement, kastner2021arena, long2018towards, tong2021uav}. Additionally, a random initialization of obstacles and agents is a common practice in training, as outlined in the introduction. However, we want to gain more control over the criticality of a collision in training. Therefore, we propose a training environment that consists of multiple short scenarios instead of long episodes with sequential obstacle encounters. Each scenario, which is equal to one training episode, is defined by the agent's initial position $\eta_0 = (0, 0, 0)^\top $, the agent's initial velocity $\nu = (u_0, 0, r_0)$, the number of obstacles $N_{\rm obst}$, the initial position $\eta_{\rm obst,0}^i = (x_{\rm obst,0}^i, y_{\rm obst,0}^i, \psi_{\rm obst,0}^i)^\top $ and velocity  $\nu_{\rm obst,0}^i = (u_{\rm obst,0}^i, 0, 0)^\top $ of all obstacles $i = 1, \dots ,N_{\rm obst}$ and the corresponding collision risk $CR$ (based on Section \ref{sec:CRmetric}). These variables, defining a scenario, are listed in Table \ref{tab:scenario_definition} with the corresponding ranges. We limit our study to a maximum of three obstacles, however, the approach can be easily scaled up.

\begin{table}[H]
    \centering
    \begin{tabular}{l|l|l}
    Scenario variable & Description & Range\\
    \toprule
     $u_0$ & agent's initial velocity  &  $[1 ,2 ] $ m/s \\
     $r_0$ & agent's initial angular velocity    &  $[-0.1 ,0.1 ]$ $s^{-1}$\\
     $x_{\rm obst,0}^i$ & initial position for obstacle $i$    &  $(-\infty, \infty)$\\
     $y_{\rm obst,0}^i$ & initial position for obstacle $i$    & $(-\infty, \infty)$\\
     $\psi_{\rm obst,0}^i$ & initial heading for obstacle $i$    & $[0 , 2\pi] $\\
     $u_{\rm obst,0}^i$ & initial velocity for obstacle $i$    &  $[0 , 2] $ m/s \\
     $N_{\rm obst}$ & number of obstacles   &  $[1 ,3] $\\
     $CR$ &  collision risk  &  $[0 ,1 ] $\\

    \end{tabular}
    \caption{Variables defining a training scenario with their corresponding range of values. }
    \label{tab:scenario_definition}
\end{table}
To control for the difficulty in training, we aim to define an environment where the $CR$ of all training scenarios follows a predefined distribution. Therefore, we need to generate scenarios based on a given collision risk. There is no explicit solution for this problem, but the following equations generally describe the dependence between the CR value and the initial agent and obstacle states, exemplary for one obstacle. On one side, the agent trajectory based on a constant action vector $a = (a_{u}, a_{r})^\top$ can be computed via integrating the agent dynamics (\ref{eq:agent_dyn}):

\begin{align*}
    x_t(a_u, a_r) &= \int_0^{t} \left(u_0 + \frac{\tau_{u,\rm max }a_u}{m} t'\right) \cos{\frac{\tau_{r,\rm max }a_r}{2 I}t'^2}dt', \\
    y_t(a_u, a_r) &= \int_0^{t} \left(u_0 + \frac{\tau_{u,\rm max }a_u}{m} t'\right) \sin{\frac{\tau_{r,\rm max }a_r}{2 I}t'^2}dt'.     
\end{align*}
The obstacle's trajectory on the other side is defined as:
\begin{align*}
    x_{{\rm obst}, t} &= x_{\rm obst,0} + u_{\rm obst,0} \cos{\psi_{\rm obst,0}}t,\\
    y_{{\rm obst}, t} &= y_{\rm obst,0} + u_{\rm obst,0} \sin{\psi_{\rm obst,0}}t.
\end{align*}
Having the distance between an agent and an obstacle as
\begin{equation*}
    D_t(a_u, a_r) = \sqrt{\left\{x_t(a_u, a_r) - x_{{\rm obst}, t}\right\}^2 
    + \left\{y_t(a_u, a_r) - y_{{\rm obst}, t}\right\}^2},
\end{equation*}
the CR can now be computed via:
\begin{align*}
    CR &= \frac{1}{N_u N_r}\sum_{i=1}^{N_u} \sum_{j=1}^{N_r}  \mathds{1}\left\{\min_t  D_t(a_{u,i}, a_{r,j}) < 2 R_{\rm coll}\right\}.
\end{align*}
where the actions $a_{u,i}$ and $a_{r,j}$ are equally spaced between $[-1, 1]$ with $N_u$ and $N_r$ defining the number of actions considered. We chose $N_u = N_r = 11$, resulting in 121 different agent trajectories.

We see that the overall function $CR(u_0, r_0, x_{\rm obst, 0},y_{\rm obst, 0},\psi_{\rm obst, 0},u_{\rm obst, 0} )$ is heavily nonlinear and that an explicit inverse with respect to any argument does not exist. For this reason, a Monte Carlo approach was used to generate a huge pool of $5\times10^6$ scenarios from which we can draw scenarios with a predefined value of $CR$. Having this pool, a custom distribution of collision risk can be achieved in training. The distributions we used in this study are defined in Section \ref{sec:main_study}.

To generate a single scenario, we first sampled $u_0, r_0, \psi_{\rm obst, 0},$ and $u_{\rm obst, 0}$ from the defined ranges in Table \ref{tab:scenario_definition}. Next, we computed $x_{\rm obst, 0}$ and $y_{\rm obst, 0}$ in a way that an obstacle collides with at least one possible agent's trajectory within the time interval  $[0.2t_{\rm lim}, t_{\rm lim}]$. This condition ensures that all obstacles pose a threat to the agent. Furthermore, the lower bound of the time interval ensures that the agent has enough time to react before a possible collision. Now that we have chosen the initial positions and velocities for the agent and the obstacles, the collision risk is computed by simulating the scenario for a varying control input $a_u$ and $a_r$. 

A training episode ends when either the agent collides with an obstacle or when the distance $d_t^i$ to each obstacle $i$ starts increasing. After an episode, the agent is getting rewarded according to (\ref{eq:reward}).
The Euler and ballistic methods \citep{Treiber2013} are used to update the velocity and position vector of the agent and obstacles at time step $t+1$. Consequently, an update for agent and obstacle $i$ is performed by:
\begin{align}
	\nu_{t+1} &= \nu_t + \Dot{\nu}_t \Delta t,\label{eq:update1}\\
	\eta_{t+1} &= \eta_t + \frac{\Dot{\eta_t}+ \Dot{\eta}_{t+1}}{2} \Delta t,\label{eq:update2}\\
	\eta_{\rm obst, t+1}^i &= \eta_{\rm obst,t}^i + \frac{\Dot{\eta}_{\rm obst,t}^i+ \Dot{\eta}_{\rm obst, t+1}^i}{2} \Delta t,\label{eq:update3}
\end{align}
with $\Delta t$ corresponding to the simulation step size.

\section{Results}\label{sec:results}
\subsection{Main study}\label{sec:main_study}
\begin{figure}[!ht]
    \centering
    \includegraphics[width=1\linewidth]{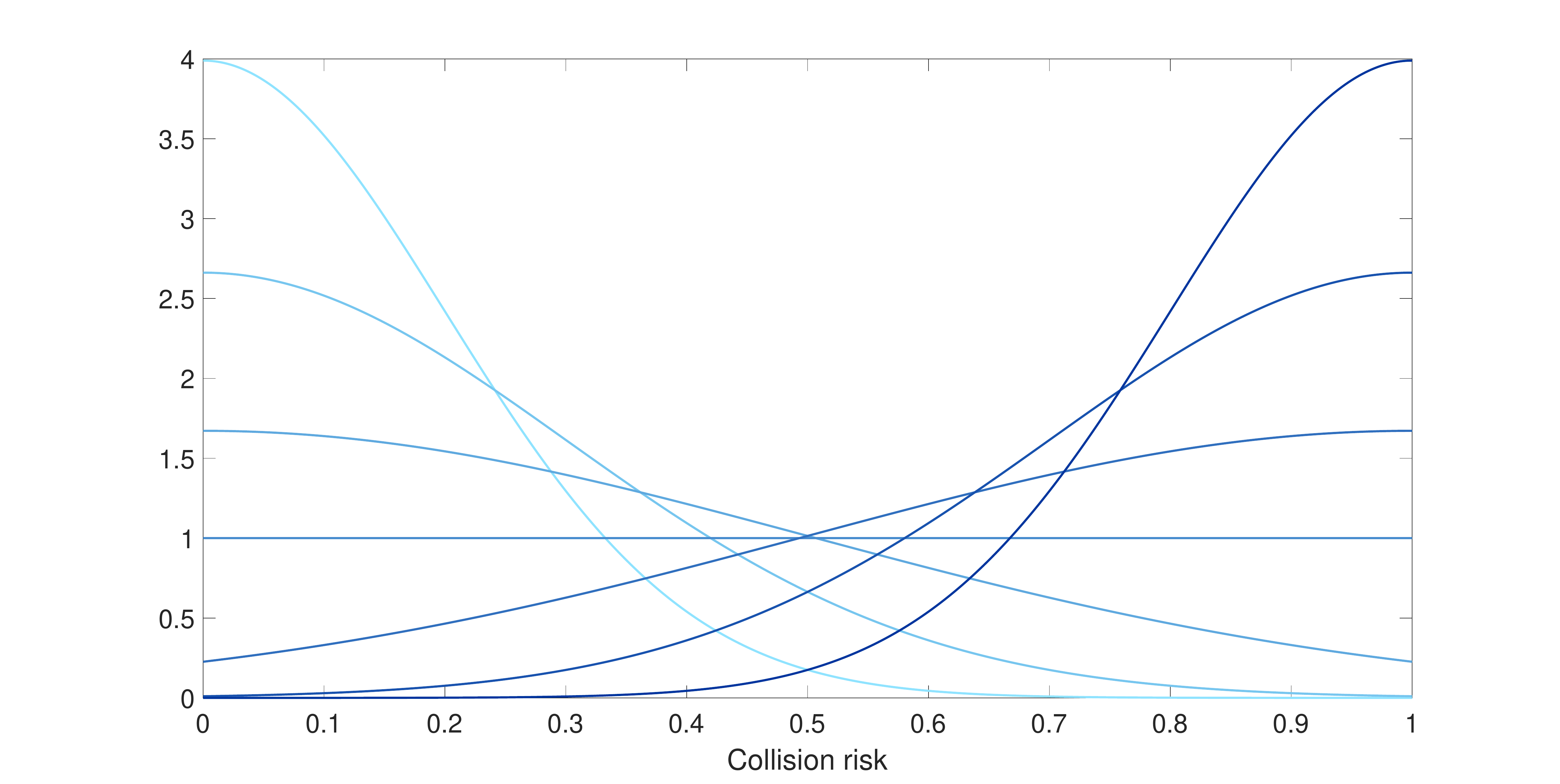}
    \caption{Seven different collision risk distributions we use in the main study, darker colors indicating distributions towards a higher collision risk.}
    \label{fig:distributions}
\end{figure}

This study investigates how the difficulty in training impacts the final performance of obstacle avoidance by adjusting two factors: First, we varied the distribution of collision risk. Figure \ref{fig:distributions} depicts the seven different distributions we used, with darker colors indicating increasing scenario difficulty according to the collision risk. And second, we varied the proportions of scenarios with one, two, and three obstacles. 
The resulting configurations are depicted in Table \ref{tab:obstacle_ratios}.

\begin{table}[H]
    \centering
    \begin{tabular}{l|l|l|l}
    Configuration & 1 obstacle & 2 obstacles & 3 obstacles\\
    \toprule
     $1/2/4$ & $14.29\%$ & $28.57 \%$ & $57.14\%$\\
     $1/1/1$ & $33.33\%$ & $33.33\%$ & $33.33\%$\\
     $4/2/1$ &  $57.14\%$ & $28.57\%$ & $14.29\%$\\
    \end{tabular}
    \caption{Three different training configurations with varying proportions of obstacles. }
    \label{tab:obstacle_ratios}
\end{table}

\begin{figure}[!ht]
    \centering
    \includegraphics[width=1\linewidth]{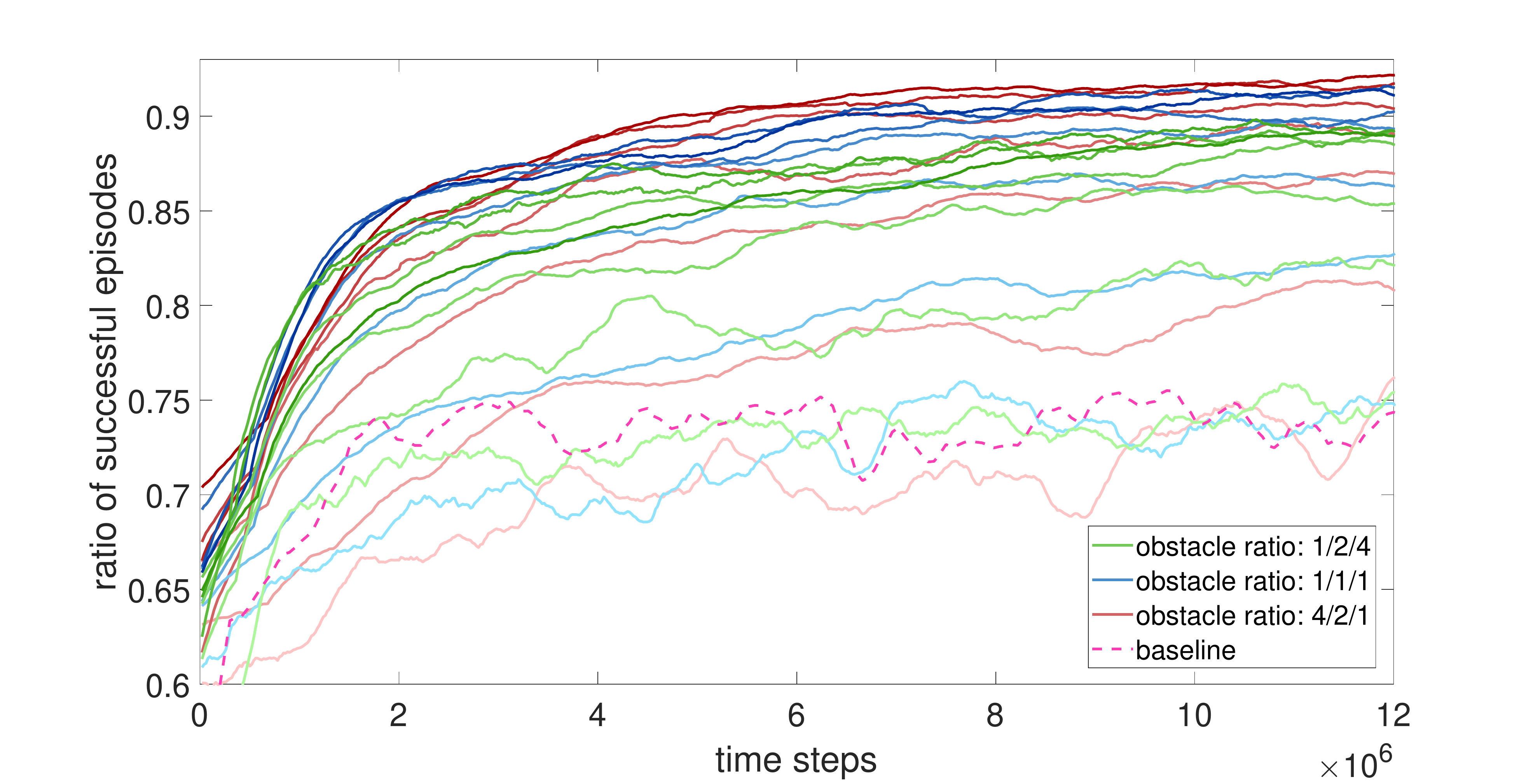}
    \caption{Training process for 21 agents, trained with all combinations of collision risk distributions (Figure \ref{fig:distributions}) and obstacle ratios (\ref{tab:obstacle_ratios}) and darker colors indicating distributions towards a higher collision risk. }
    \label{fig:TrainingPlot}
\end{figure}

Each collision risk distribution has been combined with each obstacle configuration, resulting in 21 different agents. The training process of each agent is illustrated in Figure \ref{fig:TrainingPlot}. The color coding from Figure \ref{fig:distributions} has been adopted, with darker colors indicating a higher proportion of high collision risk scenarios. The x-axis describes the number of steps in training. For every $2\times10^4$ time step, the same 2000 validation scenarios have been conducted with the current policy. These validation scenarios show equally distributed collision risk and number of obstacles and are identical for all the trained agents. The y-axis describes the ratio of successful validation scenarios. The baseline agent, depicted with a dashed line, will be discussed in Section \ref{sec:baseline}. Generally, there are two main trends to observe: The final performance increases (i) with a higher percentage of high collision risk scenarios and (ii) with more obstacles.

\begin{figure}[h]
    \centering
    \includegraphics[width=1\linewidth]{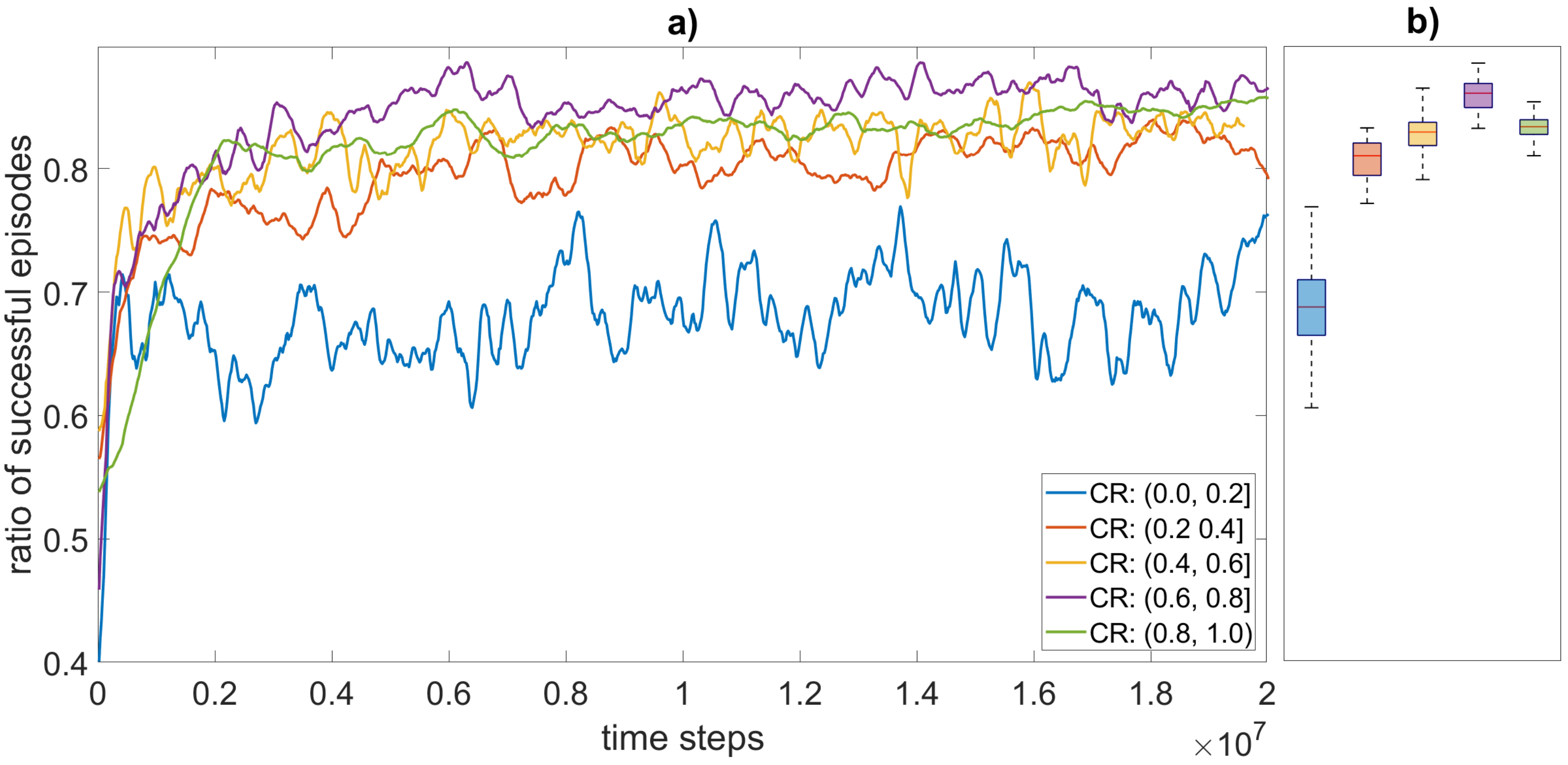}
    \caption{a) Training process for five different agents, the legend indicating the individual collision risk distribution in training. b) Distribution of performance after $2 \times 10^7$ steps of training for ten agents in each category, initialized with different seeds.}
    \label{fig:bins}
\end{figure}

Next, we investigated if we obtain even better results when the training solely covers high collision risk scenarios instead of collision risk distributions ranging between 0 and 1 (as in Figure \ref{fig:distributions}). We, therefore, trained five different agents where the collision risk was uniformly distributed in the intervals: (i) $(0, 0.2]$, (ii) $ (0.2, 0.4]$, (iii) $(0.4, 0.6]$, (iv) $(0.6, 0.8]$, (v) $(0.8, 1)$. The number of obstacles was equally distributed for all agents. Figure \ref{fig:bins} a) shows the training progress for all five configurations. Moreover, we trained ten agents in each category, initialized with different seeds. The results after $2 \times 10^7$ steps of training are depicted in Figure \ref{fig:bins} b) as boxplots.
Apparently, the training with the lowest collision risk (between 0 and 0.2) leads to a significantly lower final performance. Among the other four agents, no clear difference is observable. Eventually, slightly better performance can be assumed for the training with $CR$ values between 0.6 and 0.8. But overall, the results of the previous study can not be reached (compare with the training results of Figure \ref{fig:TrainingPlot}). Interpreting these findings, we assume that the RL agent lacks generalization abilities when just being trained with a small interval of different collision risks. Training only with high collision risks (interval $(0.8, 1)$) seems to be so complex that the agent struggles to learn satisfactorily. We, therefore, recommend defining collision risk distributions that cover all kinds of scenarios between $CR = 0$ and 1, however, strongly skewed towards higher collision risks. 
 
\subsection{Baseline agent}\label{sec:baseline}
To demonstrate the strength of the proposed approach, we trained a baseline agent in a training environment based on random initialization of the agent and obstacles and long training episodes, which is a typical approach in existing studies on RL-based obstacle avoidance. We adopted the training scheme of \cite{xu2022path} who used a quadratic box with fixed borders and five linearly moving obstacles. To ensure the obstacles are navigating within the specified area, we beam an obstacle to the other side of the box when a boundary is reached. The box itself always moves with the agent so that the agent always remains in the center of this box. Since we extended the duration of an episode to 500 time steps for training the baseline agent, we changed the reward structure to:
\begin{equation} 
\label{eq:box_reward}
	r_t = 
	\begin{cases}
		-1, & \text{when obstacle and agent are overlapping},\\
		0.01, & \text{else}.\\
	\end{cases}
\end{equation}

At the beginning of an episode, the obstacles are initialized with random directions and random velocities in the range defined in Table \ref{tab:scenario_definition}.
The validation results for the training of the baseline agent are shown in Figure \ref{fig:TrainingPlot}. We ran a hyperparameter search over the reward structure and the edge length of the quadratic box and found the reward according to (\ref{eq:box_reward}) and an edge length of $\unit[35]{m}$ to be optimal. Still, the baseline agent masters just around  $70-75\%$ of validation scenarios that were identical to the collision risk-based training. 

\begin{figure}[h]
    \centering
    \includegraphics[width=1\linewidth]{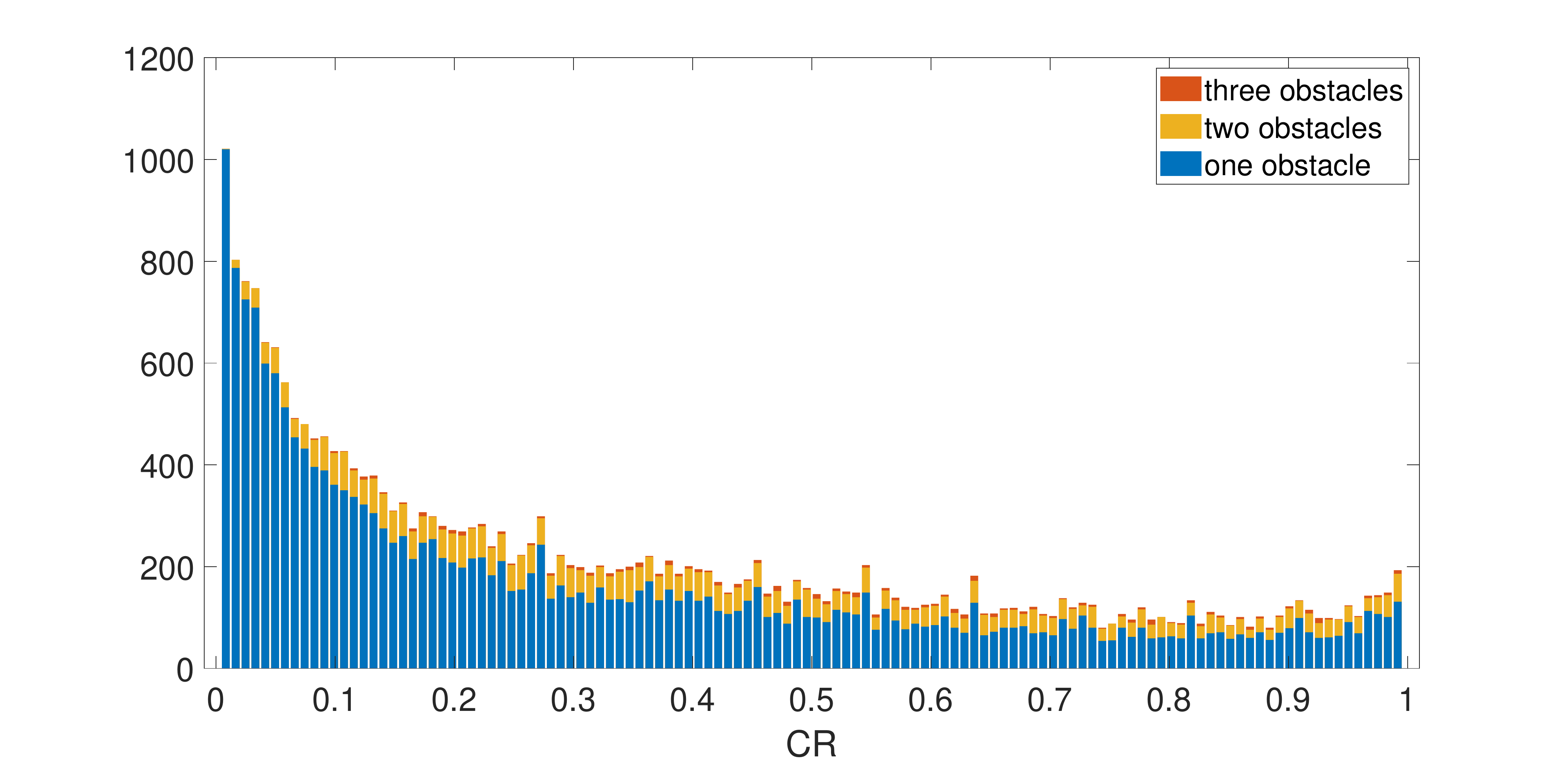}
    \caption{Distribution of collision risk in the baseline training approach, the colors indicating a different number of obstacles involved.}
    \label{fig:cr_box}
\end{figure}

To gain more insights into the weaknesses of the baseline training approach, we measured the CR distribution and the number of involved obstacles for a random initialization. We took a sample of 64 000 scenarios and found that roughly $\unit[36]{\%}$ of scenarios had zero collision risk according to the collision risk metric defined in Section \ref{sec:CRmetric}, and $\unit[24]{\%}$ showing a collision risk equal to one, meaning that a collision is unavoidable. Figure \ref{fig:cr_box} illustrates the collision risk distribution for the remaining scenarios, different colors indicating a different number of involved obstacles. We observe that this training approach suffers from (i) a high coverage of scenarios with zero collision risk, (ii) a high coverage of scenarios involving just one obstacle (blue bars), and the distribution shifted towards lower collision risk, (iii) a low coverage of more difficult scenarios including two or three obstacles (yellow and red bars), and (iv) a high coverage of scenarios where the agent has no chance to avoid the obstacle ($CR = 1$). We assume that this unfavorable configuration in training leads to the significantly lower performance of the baseline agent, seen in Figure \ref{fig:TrainingPlot}.

\subsection{Scenario-based validation}
\begin{figure}[!ht]
    \centering
    \makebox[\textwidth][c]{\includegraphics[width=1\textwidth]{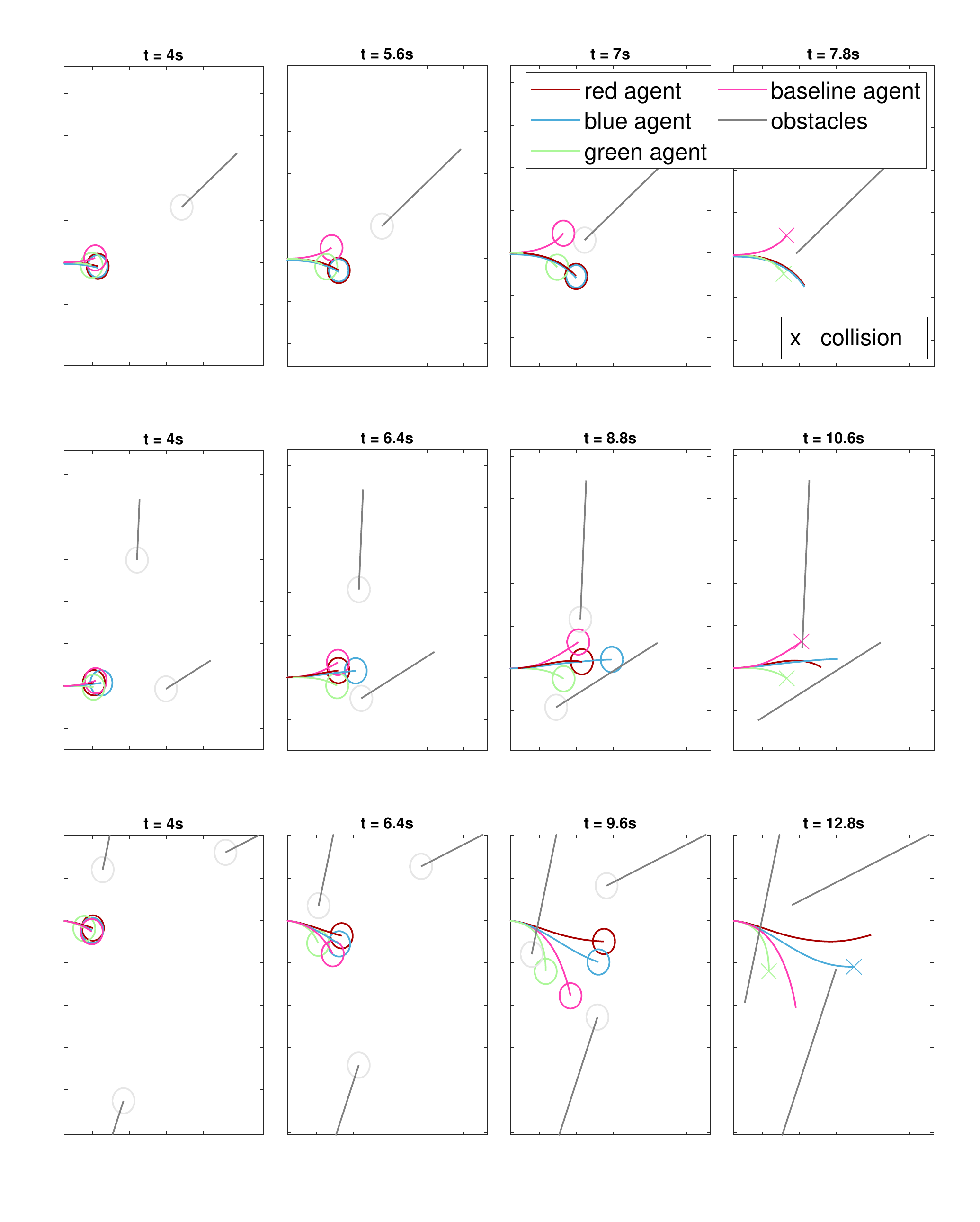}}
    \caption{Validation of our proposed approach in three scenarios with one (top row), two (middle row), and three (bottom row) obstacles by comparison of trajectories at different points in time. }
    \label{fig:ValidationScenarios}
\end{figure}

To further validate our proposed approach, we take a closer look at the trajectories of different trained agents. Figure \ref{fig:ValidationScenarios} illustrates three scenarios, each at four different points in time. The first three time frames show the agent's and obstacles' collision area, while the fourth time frame illustrates the agent's location of collision with an obstacle marked by a cross. We simulated four agents, their colors being identical to the colors in the training plot from Figure \ref{fig:TrainingPlot}. Among them, (i) the agent trained with an obstacle ratio of  $1/2/4$ (compare with Table \ref{tab:obstacle_ratios}) and the most extreme collision risk distribution (compare with Figure \ref{fig:distributions}), labeled as 'red agent', (ii) the agent trained with an obstacle ratio of $1/1/1$ and a uniform distribution of collision risk ('blue agent'), (iii) the agent trained with an obstacle ratio of $4/2/1$ and the lowest collision risk distribution ('green agent'), and (iv) the baseline agent ('pink agent'). Obstacles are marked with grey.
\clearpage

The first scenario (top row) shows one threatening obstacle and the reaction of the four defined agents. While the red, blue, and green agents try to avoid the obstacle by moving downwards, the pink agent makes the initial decision to move up, resulting in a collision with the obstacle at $t=\unit[7]{s}$. By making the wrong decision at the start of the episode, we assume that this agent lacks the ability to correctly anticipate the trajectory of the obstacle. Although the green agent decides to move down, it suffers from enough acceleration, leading to a collision at $t = \unit[7.8]{s}$. Only the red and blue agents are able to simultaneously control both actions, torque and acceleration, in a way that the task is mastered successfully.

The second scenario consists of two threatening obstacles, the upper one forcing the agent to move down, however, the second obstacle blocking this way to escape. The pink agent seems to only consider the closer obstacle (lower one) and decides to move up, however, resulting in a collision at $t = \unit[8.8]{s}$ with the upper obstacle. The green agent also seems unable to consider both obstacles simultaneously and decides to move down, resulting in a collision shortly after $t = \unit6.4{s}$. In contrast, the blue and red agents are able to correctly estimate the trajectory of both obstacles and to react accordingly. The blue agent successfully finds his way through the obstacles, yet not far away from a collision at $t = \unit[6.4]{s}$. The red agent masters the scenario as well, with enough safety distance to the lower obstacle, however, also closely missing a collision at $t = \unit[10.6]{s}$.

In a last scenario (bottom row), the agent is facing three threatening obstacles. The green and pink agents again seem to consider just the closest and most threatening obstacle, and both react by moving downwards, the green one colliding with an obstacle at $t = \unit[9.6]{s}$. On the contrary, red and blue agents seem to be able to include all three obstacles in the decision of maneuver planning, resulting in a more complex, "S"-shaped trajectory. However, among them, the red agent is the only one without collision. 

Looking at all scenarios together, the red agent seems to know best how to predict future obstacle trajectories and how to avoid these threats under consideration of its own maneuverability. Especially in the last scenario, this agent shows its capability to do complex collision avoidance maneuvers under consideration of multiple obstacles at once. 

\section{Generalizability study}\label{sec:generelizability}
After demonstrating the strength of the proposed training approach for specific agent dynamics, we will investigate how generally this approach can be applied to different domains. We, therefore, adopted the proposed training method in two common dynamic obstacle avoidance settings: First, in RL-based dynamic obstacle avoidance for mobile robots (Section \ref{sec:mobile_robot}), which, for example, has been studied in \cite{huang2018reinforcement}, \cite{arvind2019autonomous}, \cite{everett2018motion}, or \cite{fan2020distributed}; and second, in RL-based maritime ship-collision avoidance (Section \ref{sec:ship}), which is a common field of scientific research as well \citep{zhao2019colregs, XU2020, xie2020composite, chun2021deep}. Therefore, we defined two further agent dynamics that both built upon the general definition of a 3-DoF model (equations (\ref{eq:nu_dot}) and (\ref{eq:eta_dot})) and can be interpreted as an extension of the model that we used in the previous study. The training results for both use cases are depicted in Section \ref{sec:generalizability_results}. 
Another part of the generalizability study is to analyze the robustness of the proposed training approach when experiencing sensor signal noise, detailed in Section \ref{sec:noise}.

\subsection{Mobile robot collision avoidance}\label{sec:mobile_robot}
\begin{figure}[!ht]
    \centering
    \includegraphics[width=0.7\linewidth]{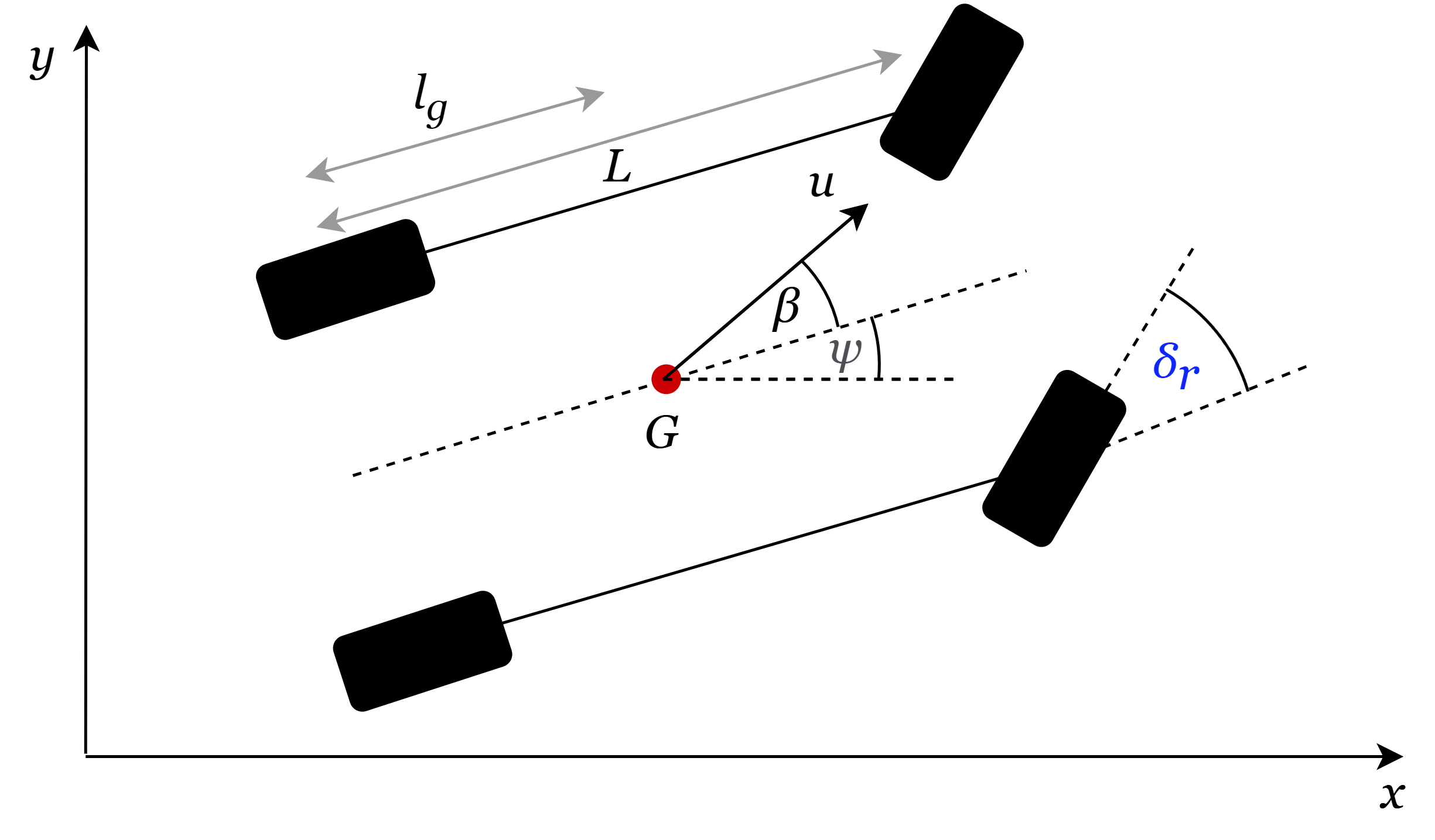}
    \caption{Technical drawing of a four-wheeled mobile robot with steering of the front axis.}
    \label{fig:robot}
\end{figure}

We use a two-dimensional model of a four-wheeled robot with control of the acceleration force $\tau_u$ and the steering angle of the front axis $\delta_r$. For a schematic illustration, see Figure \ref{fig:robot}. The resulting kinematic model can then be written as \citep{rajamani2011vehicle}:
\begin{align*}
    \Dot{x} &= u \cos(\psi + \beta),\\
    \Dot{y} &= u \sin(\psi + \beta),\\
    \Dot{\psi} &= \frac{u}{l_r}\sin(\beta),\\
    \Dot{u} &= \frac{\tau_u}{m},\\
    \Dot{v} &= 0,
\end{align*}
where $\beta$ is the slip angle at the center of gravity $G$:
\begin{align*}
    \beta &= \arctan\left\{\tan(\delta_r)\frac{l_r}{L}\right\}.
\end{align*}

As in the main study, we configured two-dimensional action $a_t \in [-1, 1]^2$ that is linearly mapped to the acceleration force and the steering angle by using a maximum acceleration force  $\tau_{u,\rm max}$ and a maximum steering angle $\delta_{r,\rm max}$. We further adopted the state space (\ref{eq:state_space}) and the update scheme (\ref{eq:update1}) - (\ref{eq:update3}) from the main study. To use the same training approach, the scaling parameters, as well as the ranges of scenario variables, had to be adjusted. All robot and environment related parameters and variable ranges are depicted in Table \ref{tab:robot_parameters}. 

\subsection{Maritime ship collision avoidance}\label{sec:ship}
Next to collision avoidance for a mobile robot, we aim to prove our proposed training approach for obstacle avoidance in a common maritime traffic application. Therefore, we consider the widely used KVLCC2 tanker with a length $L = \unit[320]{m}$ and adopt the 3-DoF ship maneuvering model from \cite{yasukawa2015introduction}. Again, we use $\eta = (x, y, \psi)^\top  \in	\mathbb{R}^3 $ and $\nu = (u, v, r)^\top  \in	\mathbb{R}^3$ to describe position and velocity vector. The ship dynamics follows the general 3-DoF model (\ref{eq:nu_dot}) and (\ref{eq:eta_dot}) as well, however, the resulting differential equations can not be expressed explicitly:
\begin{align*}
\left(m+m_{x}\right) \dot{u}-\left(m+m_{y}\right) v r-x_{G} m r^{2} &=X, \\
\left(m+m_{y}\right) \dot{v}+\left(m+m_{x}\right) u r+x_{G} m \dot{r} &=Y, \\
\left(I_{z G}+x_{G}^{2} m+J_{z}\right) \dot{r}+x_{G} m(\dot{v}+u r) &=N_{m},
\end{align*}
with ship mass $m$, added masses $m_x$ and $m_y$ in $x$ and $y$ direction respectively, the longitudinal coordinate of the center of gravity  $x_G$, the moment of inertia  $I_{zG}$ of the ship around the center of gravity, and the added moment of inertia $J_z$. The surge force $X$, lateral force $Y$, and the yaw moment $N_m$ around midship can be decomposed in components acting on hull (H), rudder (R), and propeller (P):
\begin{align*}
X &= X_{H}+X_{R}+X_{P}, \\
Y &=Y_{H}+Y_{R}, \\
N_{m}&=N_{H}+N_{R}.
\end{align*}
The ship motion can be controlled by rudder angle $\delta_r$ and propeller revolutions $n_{\rm P}$ which directly impact surge force $X$ lateral force $Y$, and the yaw moment $N_m$ around midship. We refer to \cite{yasukawa2015introduction} for a detailed explanation of each force component and the derivation of parameters describing the KVLCC2 tanker. 

Again, we configured two-dimensional action $a_t \in [-1, 1]^2$ that is linearly mapped to propeller revolutions $n_{\rm P} \in [0, n_{\rm P,\rm max}]$ and the rudder angle $\delta_r \in [-\delta_{r, \rm max}, \delta_{r, \rm max}]$. Since, in contrast to the previously used models, the ship model is not restricted to lateral velocities $v = 0$, the state space from  (\ref{eq:state_space}) is extended by the factor $v_t/u_{\rm scale}$. All environment-related parameters with corresponding values and ranges are depicted in Table \ref{tab:robot_parameters}.

\subsection{Training results}\label{sec:generalizability_results}
\begin{figure}[!ht]
    \centering
    \includegraphics[width=1\linewidth]{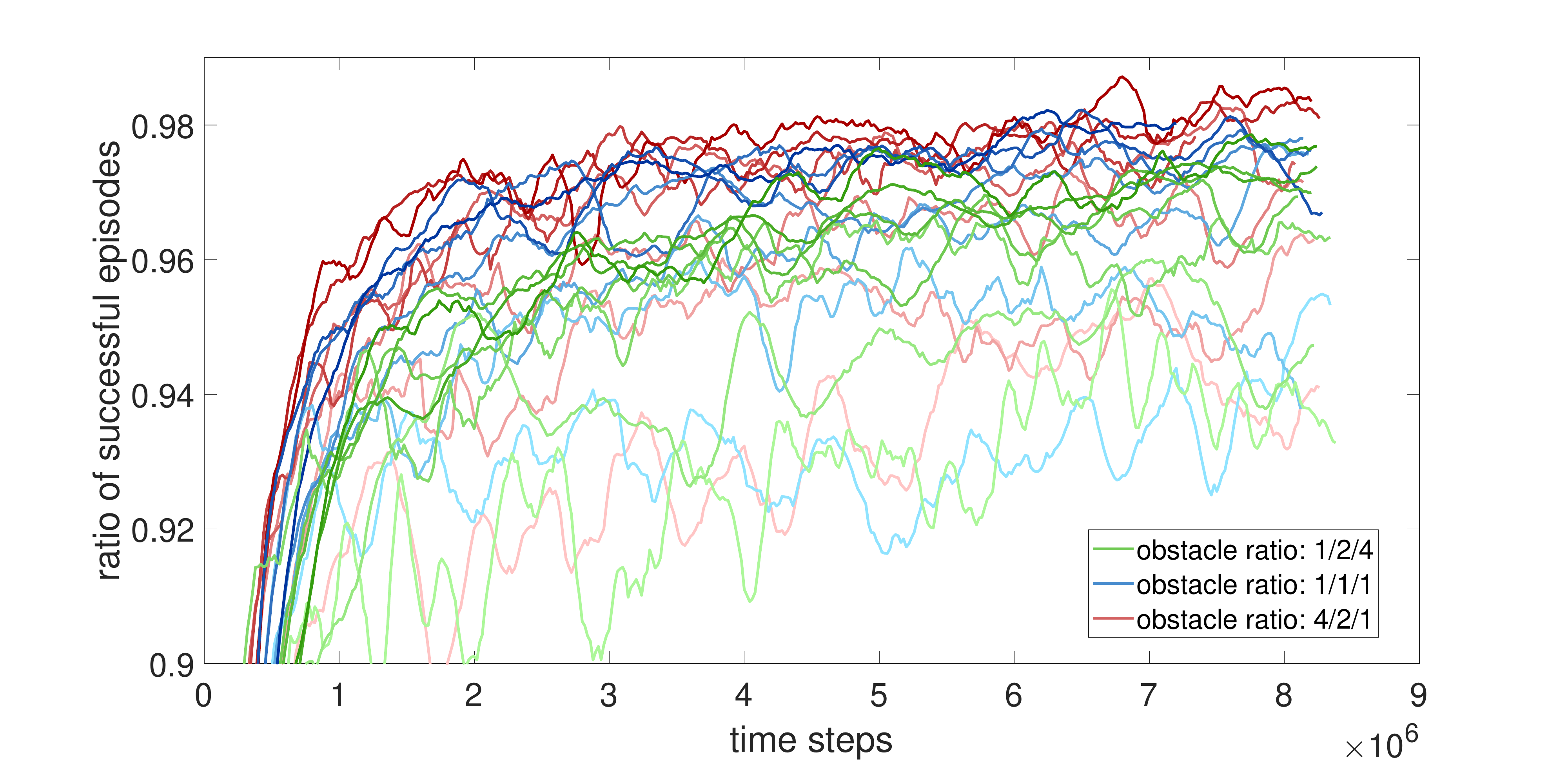}
    \caption{Training process of the mobile robot in the proposed training environment, darker colors indicating a higher collision risk in training.}
    \label{fig:trainingPlot_robot}
\end{figure}

\begin{figure}[!ht]
    \centering
    \includegraphics[width=1\linewidth]{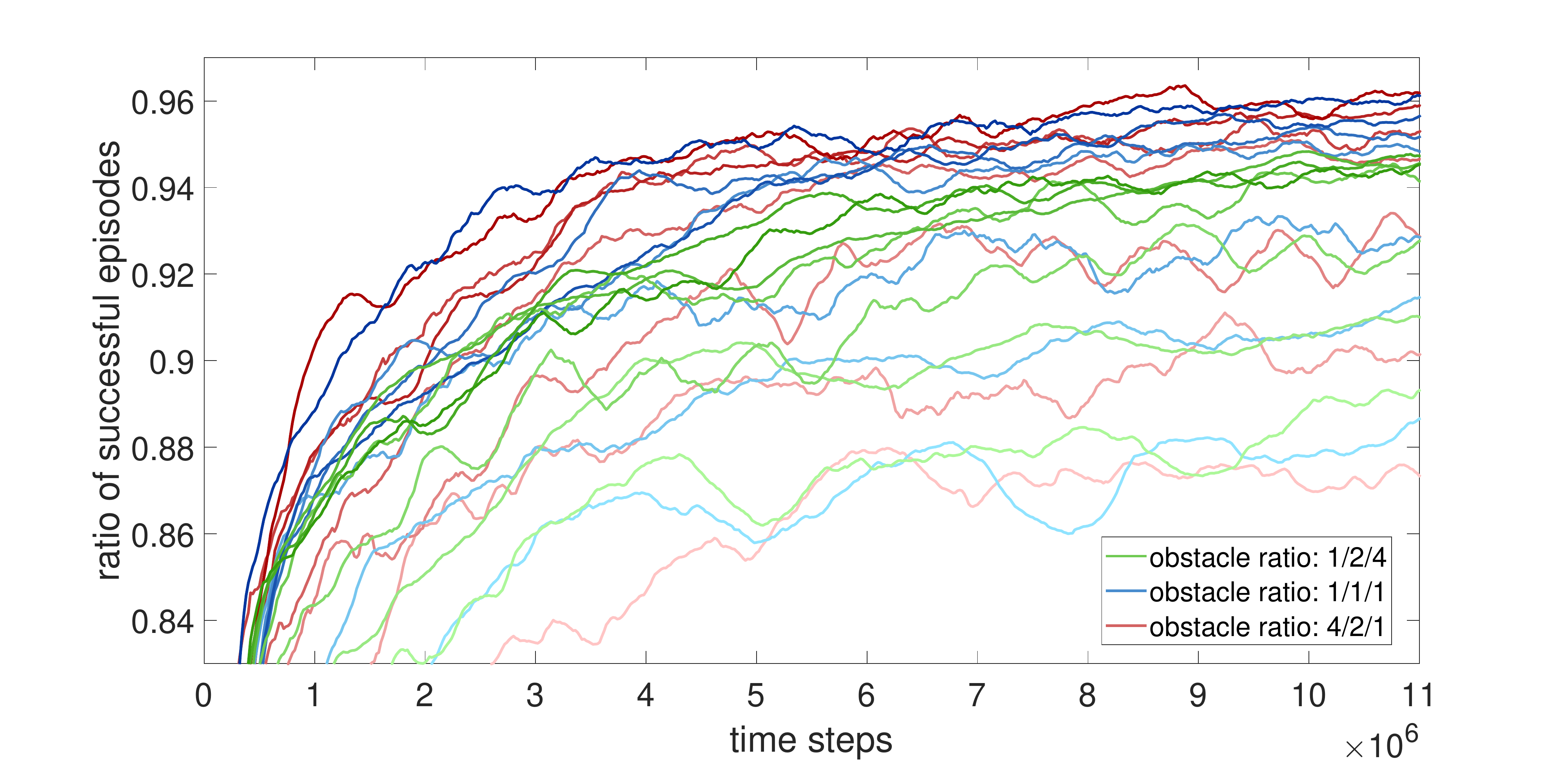}
    \caption{Training process of the KVLCC2 tanker in the proposed training environment.}
    \label{fig:trainingPlot_ship}
\end{figure}

We conducted two further trainings based on the previously defined agent dynamics. To evaluate the generalizability of our proposed approach, we used the same obstacle and collision risk distribution as in the main study (see Figure \ref{fig:distributions} and  Table \ref{tab:obstacle_ratios}). Figure \ref{fig:trainingPlot_robot} and \ref{fig:trainingPlot_ship} depict the training process based on the robot and the ship dynamics, respectively. We used the same color coding as in the main study for obstacle and collision risk distributions. The general trend we observed before can be confirmed: Obstacles ratios towards more obstacles and collision risk distributions towards higher risks lead to a higher ratio of successful validation episodes. These findings show that instead of being limited to a specific problem, our training approach can be applied to many different domains of obstacle avoidance. 

\subsection{Sensor noise}\label{sec:noise}
\begin{figure}[!ht]
    \centering
    \includegraphics[width=1\linewidth]{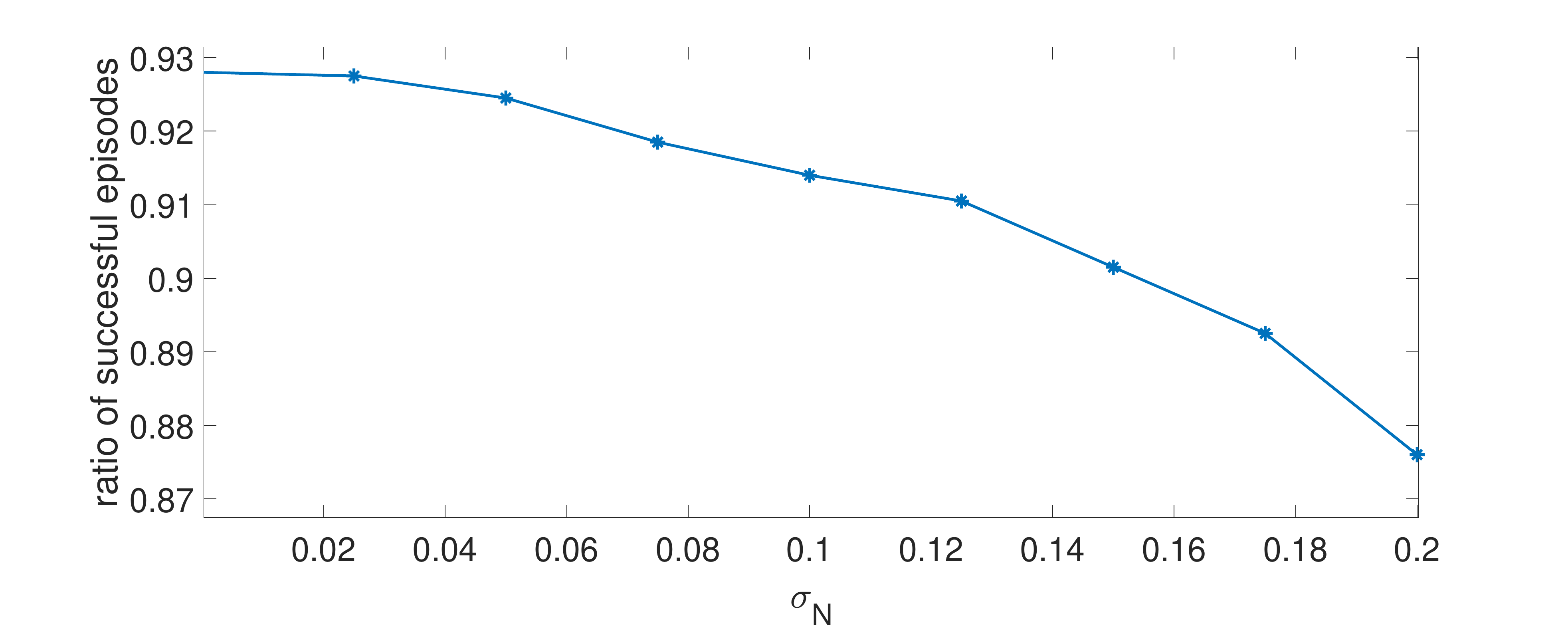}
    \caption{Robustness of the trained agent against Gaussian sensor noise with standard deviation $\sigma_{\rm N}$.}
    \label{fig:noise}
\end{figure}

In another experiment, we aim to analyze the robustness of the training approach against noisy sensor signals. Therefore, Gaussian noise is multiplied to all state variables (\ref{eq:state_space}) each time step. For an exemplary state variable $s^*$, this results in:

\begin{equation*}
    s_{\rm N}^* = s^* \epsilon, \quad \text{with } \epsilon \sim \mathcal{N}(1, \sigma_{\rm N}),
\end{equation*}
with standard deviation $\sigma_{\rm N}$. Figure \ref{fig:noise} depicts the ratio of successful validation episodes for varying noise intensity. The results show that even for quite noisy signals with a standard deviation of $\sigma_{\rm N} = 0.15$, the trained agent is still able to master a majority of the validation scenarios successfully. This demonstrates the robustness of the proposed training approach in situations where sensory information like positions and velocities of obstacles can not accurately be measured.

\section{Conclusion}\label{sec:conclusion}
This RL-based study analyzed the effect of varying difficulty in training on the final performance of obstacle avoidance. We hypothesized that resulting from a random initialization of agent and obstacles and long training episodes, usual training approaches suffer from a low coverage of high-risk scenarios, leading to an impaired final performance. To overcome this issue, we proposed a scenario-based training environment where the difficulty of a scenario is controlled. We assessed the difficulty of a training scenario by the number of threatening obstacles and an existing collision risk metric. We defined different ratios of obstacles and multiple distributions of collision risk to define scenarios ranging from simple obstacle avoidance with one obstacle and a low probability of collision up to high collision risk scenarios with three threatening obstacles. The training of an initial agent with simple dynamics indicates that training scenarios covering more obstacles and higher initial collision probabilities lead to better final performance of obstacle avoidance. 

For better comparability of our proposed approach, we defined a baseline training environment with random initialization of agent and obstacles and long training episodes, which is a common method in different applications of dynamic obstacle avoidance. The poor results of an agent trained in the defined baseline environment confirm our approach, and an analysis of collision risk distribution in training indicates that a random initialization of agent and obstacles suffers from a low coverage of situations with a high probability of collision. 

Another feature of our proposed method is its suitability for different application domains of obstacle avoidance, regardless of the agent's maneuverability. This advantage results from the used collision risk metric that explicitly considers the agent's dynamics. We proved the generalizability in two common applications of dynamic obstacle avoidance: A mobile robot and a maritime ship under the threat of approaching obstacles that can be considered as other traffic participants. In both applications, we can confirm the advantage of using training scenarios shifted towards greater difficulty. To further prove the robustness of the trained agent, we added Gaussian noise to the sensor signal, resulting in just a marginal degradation of performance.

However, there are still several issues that need to be addressed. Instead of using the number of threatening obstacles and the adopted collision risk metric, one should think about developing more sophisticated methods to assess the collision probability in a training scenario. A disadvantage coming along with the collision risk metric from \cite{huang2020time} is the assumption of a constant control vector to compute the collision probability. Another assumption is the constancy of obstacle trajectories which further restricts our proposed approach. 

Despite those drawbacks and generalizing our proposed approach with regard to other obstacle avoidance applications, we believe that employing scenario-based training with increased coverage of high collision risk scenarios, whatever collision risk metric might be used, generally results in a better final performance.

\subsubsection*{Acknowledgements}
This work was partially funded by BAW - Bundesanstalt für Wasserbau (Mikrosimulation des Schiffsverkehrs auf dem Niederrhein).

\newpage
\bibliographystyle{elsarticle-harv} 
\bibliography{elsarticle-template-harv}

\newpage
\appendix
\setcounter{table}{0}
\renewcommand{\thetable}{A.\arabic{table}}

\begin{algorithm}[h]
\setstretch{1.05}
\SetAlgoLined
 Randomly initialize critics $Q^{\omega_1}, Q^{\omega_2}$ and actor $\mu^{\theta}$\\
 Initialize target critics $Q^{\omega'_1}, Q^{\omega'_2}$ and target actor $\mu^{\theta'}$ with $\omega'_1 \leftarrow \omega_1$, $\omega'_2 \leftarrow \omega_2$, $\theta^{'} \leftarrow \theta$\\
 Initialize replay buffer $\mathcal{D}$\\
 Receive initial state $s_1$ from environment \\
 \For{t = 1,T}{
 \emph{Acting}\\
 Select action with exploration noise: $a_t = \mu^{\theta}(s_t) + \epsilon$, \quad $\epsilon \sim \mathcal{N}(0, \sigma)$\\
 Execute $a_t$, receive reward $r_{t+1}$, new state $s_{t+1}$, and done flag $d_t$\\
 Store transition $(s_t, a_t, r_{t+1}, s_{t+1}, d_t)$ to $\mathcal{D}$\\
 \medskip
 \emph{Learning}\\
  Sample random mini-batch of transitions $(s_i, a_i, r_{i+1}, s_{i+1}, d_i)_{i=1}^{N} $ from $\mathcal{D}$\\
  Calculate targets:\\
  \vspace{-1cm}
  \begin{align*}
      \Tilde{a}_{i+1} &= \mu^{\theta'}(s_{i+1}) + \Tilde{\epsilon}, \quad \Tilde{\epsilon} \sim \text{clip}\{\mathcal{N}(0, \Tilde{\sigma}),-c,c\},\\
      y_i &= r_{i+1} + \gamma (1-d_i) \min_{j=1,2} Q^{\omega'_j}(s_{i+1},\Tilde{a}_{i+1}).
  \end{align*} \\
  \vspace{-0.5cm}
 Update critics: $\omega_j \leftarrow \min_{\omega_j}N^{-1} \sum_{i=1}^{N} \left\{y_i - Q^{\omega_j}(s_i, a_i)\right\}^2$\\
 \If{$t \mod d$}{
 Update actor: $\theta \leftarrow \max_{\theta} N^{-1} \sum_{i=1}^{N} Q^{\omega_1}\left\{s_i, \mu^{\theta}(s_i)\right\}$\\
  Update target networks via (\ref{eq:DDPG_soft_tgt_up})
 }
 \medskip
\emph{End of episode handling}\\
    \If{$d_t$}{
    Reset environment to an initial state $s_{t+1}$\\
    }
}
 \caption{TD3 algorithm following \cite{fujimoto2018addressing}.}
 \label{algo:TD3}
\end{algorithm}

\begin{table}[H]
    \centering
    \begin{tabular}{l l| l l}
  \multicolumn{2}{c}{Mobile robot}&  \multicolumn{2}{c}{Maritime ship} \\
    \toprule
    $m$ & $\unit[5]{kg}$ & \multicolumn{2}{c}{-}\\
    $ l_r $ & $\unit[1]{m}$ &\multicolumn{2}{c}{-} \\
    $ L $ & $\unit[2]{m}$  & \multicolumn{2}{c}{-}\\
    $R_{\rm coll}$ & $\unit[6]{m}$ &  $R_{\rm coll}$ & $\unit[300]{m}$ \\
    $\tau_{u, \max}$  & $\unit[1]{N}$  & $n_{\rm P, \max}$  & $\unit[2]{s^{-1}}$ \\
    $\delta_{r, \max}$  & $\unit[5]{^{\circ}}$  &$\delta_{r, \max}$  & $\unit[20]{^{\circ}}$\\
    $u_{\rm scale}$ & $\unit[6]{m/s}$  & $u_{\rm scale}$ & $\unit[8]{m/s}$\\
    $r_{\rm scale}$ & $\unit[0.2]{s^{-1}}$  & $r_{\rm scale}$ & $\unit[0.005]{s^{-1}}$\\
    $d_{\rm scale}$ & $\unit[40]{m}$  &  $d_{\rm scale}$ & $\unit[2000]{m}$  \\
    $t_{\rm scale}$ & $\unit[20]{s}$  &  $t_{\rm scale}$ & $\unit[1400]{s}$ \\
    $\Delta t$ & $\unit[0.1]{s}$  & $\Delta t$ & $\unit[7]{s}$ \\
       $u_{\rm obst,0}^i$ & $[0 , 4] $ m/s  &  $u_{\rm obst,0}^i$ & $[0 , 8] $ m/s \\
        $u_0$ &  $[2 ,4 ]$ m/s    &$u_0$ &  $[4 ,8 ]$ m/s \\
    \multicolumn{2}{c}{-}   & $ r_0 $ & $[-4, 4] $ $10^{-3}\rm s^{-1}$   \\

    \end{tabular}
    \caption{Mobile robot and maritime ship environment parameters and variable ranges.}
    \label{tab:robot_parameters}
\end{table}



\end{document}